# Exploring Post-Training Alignment of Small Language Models for Biomedical Data-to-Text Generation: A Case Study of Medication Leaflet

Xi Yang, PhD
*Advanced Intelligence and Research*
*Eli Lilly and Company*
Indianapolis, IN, USA

Guodong Liu, PhD
*Advanced Intelligence and Research*
*Eli Lilly and Company*
Indianapolis, IN, USA

Chuqin Li, PhD
*Advanced Intelligence and Research*
*Eli Lilly and Company*
Indianapolis, IN, USA

Fan Wu, PhD
*Advanced Intelligence and Research*
*Eli Lilly and Company*
Indianapolis, IN, USA

Ergin Soysal, PhD
*Advanced Intelligence and Research*
*Eli Lilly and Company*
Indianapolis, IN, USA

Min Jiang, PhD
*Advanced Intelligence and Research*
*Eli Lilly and Company*
Indianapolis, IN, USA

Xing He, PhD
*Dept. of Biostatistics and Health Data Science*
*Indiana University*
Indianapolis, IN, USA

Jiang Bian, PhD
*Dept. of Biostatistics and Health Data Science*
*Indiana University*
Indianapolis, IN, USA

Yi Guo, PhD
*Dept. of Health Outcomes and Biomedical Informatics*
*University of Florida*
Gainesville, FL, USA

Shams Zaman, PhD
*Advanced Intelligence and Research*
*Eli Lilly and Company*
Indianapolis, IN, USA

Thomas Fuchs, PhD
*Advanced Intelligence and Research*
*Eli Lilly and Company*
Indianapolis, IN, USA

Yonghui Wu, PhD
*Dept. of Health Outcomes and Biomedical Informatics*
*University of Florida*
Gainesville, FL, USA

***Abstract*—Translating structured biomedical information into accurate, patient-friendly narratives remains a challenging biomedical data-to-text (D2T) task. While small language models (SLMs) are attractive for healthcare applications because of their lower computational cost, easier deployment, and improved data governance, the impact of post-training alignment on biomedical D2T generation has not been systematically studied. In this work, we compare four alignment methods—supervised fine-tuning (SFT), direct preference optimization (DPO), odds ratio preference optimization (ORPO), and group relative policy optimization (GRPO)—using Qwen-2.5 models for medication leaflet generation. We further introduce FDALeaflets, a dataset derived from OpenFDA drug labels, to evaluate cross-regulatory generalization in a zero-shot setting. Selected aligned SLMs outperform GPT-5 on several reference-based metrics, whereas GPT-5 achieves better hallucination and adequacy scores on FDALeaflets under LLM-as-judge evaluation. Among RL-free alignment methods, ORPO achieves the best in-domain semantic performance, reaching a semantic textual similarity (STS) score of 0.9752. GRPO achieves the highest FDALeaflets STS (0.9407), exceeding GPT-5 (0.9295). These findings show that post-training alignment is a key factor in improving both semantic fidelity and generalization for biomedical D2T generation, while clinical safety and patient readability require human verification.**

## I. Introduction

Medication leaflets are a central component of patient-facing communication, translating regulatory drug information into accessible narratives that describe indications, dosing, contraindications, and potential adverse events. [1], [2] Regulatory agencies such as the European Medicines Agency (EMA) and the U.S. Food and Drug Administration (FDA) publish drug information in highly structured, machine-readable formats (e.g., FDA structured product labeling – SPL) to support regulatory review and compliance. [3], [4] However, transforming these standardized inputs into clear, readable, and patient-appropriate narrative leaflets remains largely a manual process that is costly, labor-intensive, and difficult to scale. [5] From a biomedical informatics perspective, medication leaflet generation can be framed as a data-to-text (D2T) generation task: converting structured regulatory data into semantically faithful natural-language patient-facing explanations that preserve clinical accuracy and completeness, comply with regulatory and safety requirements, but also meet readability expectations. [6], [7]

Advances in pretrained language models have substantially improved biomedical natural language processing (NLP) tasks, including information extraction, summarization, question answering, and clinical narratives generation. [8] Related radiology research emphasizes patient-friendly reporting and evaluation beyond lexical overlap. [9] Within this landscape, small language models (SLMs) – typically defined as having <10 billion parameters – have emerged as particularly attractive for regulated healthcare settings due to their lower computational requirements, feasibility of on-premises deployment, and greater potential for auditability and governance. [10] These characteristics align closely with institutional constraints around privacy, reproducibility, and long-term model validation. Despite growing interest in applying language models for biomedical tasks, the use of SLMs for medication leaflet generation remains underexplored. It is unclear how different post-training alignment strategies affect semantic fidelity in structured biomedical D2T tasks, and whether models trained within one regulatory context can generalize to others without retraining.

Prior work in this domain is limited. To our knowledge, the BioLeaflets study represents the only published effort that explicitly investigates neural generation of medication leaflets from structured data. [7] That work introduced the BioLeaflets dataset, curated from EMA drug information, and evaluated encoder–decoder transformer models (T5 and BART) trained via supervised fine-tuning (SFT). While it established an important proof of concept, it did not compare against proprietary large language model baselines (e.g., GPT-4 or GPT-5), did not examine more recent decoder-only SLM architectures, and did not explore post-training alignment methods beyond standard supervised learning. Moreover, evaluation was limited to a single dataset, leaving open questions regarding robustness and generalizability under regulatory (e.g., EMA vs. FDA) and structural distribution shifts.

In this study, we extend prior work by systematically evaluating Qwen-2.5-series SLMs [11] for medication leaflet generation from structured regulatory inputs. Using BioLeaflets as the primary training and in-distribution evaluation dataset, we compare multiple post-training alignment strategies, including supervised fine-tuning (SFT), [12] direct preference optimization (DPO), [13] odds ratio preference optimization (ORPO), [14] and group relative policy optimization (GRPO), [15] across model scales and experimental settings. To examine generalizability, we introduce FDALeaflets, a dataset derived from OpenFDA drug labels [16] that reflects distinct regulatory conventions, section taxonomies, and linguistic patterns (compared to BioLeaflets curated from EMA drug information). We evaluate how models trained exclusively on EMA-aligned data perform under this out-of-distribution FDA setting and compare their performance against a proprietary large language model baseline (GPT-5). By jointly analyzing alignment strategies and cross-regulatory generalization, this work contributes empirical evidence on how modern SLM training approaches influence semantic reliability and robustness in patient-facing biomedical D2T generation. From a biomedical informatics perspective, these findings inform the design of scalable, governable, and regulation-aware language technologies with direct applicability to real-world clinical communication workflows.

## II. Methods

### A. Dataset

This study uses a D2T dataset focusing on medication leaflets – BioLeaflets curated from a corpus of 1,336 package leaflets of medicines sampled from the European Medicines Agency website. To prepare datasets for training and evaluation, each leaflet is split into six sections, including product overview, pre-administration information, dosage instructions, side effects, storage instructions, and package & other information identified by heuristic rules. Duplicated sections are removed. Whole leaflets are randomly assigned to training or test sets, keeping their sections together and yielding 6,640 training and 742 test samples.

To test model generalizability, we construct a new drug label dataset using drug labeling information collected from OpenFDA, which consists of the most recent drug listing information (i.e., drug labels and other drug-specific information) submitted to the U.S. Food and Drug Administration (FDA) by pharmaceutical companies. In the raw data, the drug labels are split into sections such as indication and usage, adverse reactions, and purpose, etc. Due to the characteristics of each drug product, the total variation of sections is over 100 among drug products. To align with the BioLeaflets dataset, we select 4 common section types, including "indications_and_usage", "dosage_and_administration", "how_supplied", and "adverse_reactions". For each type of section, we randomly collect 100 samples from the raw data. The selected FDA texts serve as reference outputs. Qwen3-235B-Thinking extracts structured inputs from these texts to create FDALeaflets. We use BioLeaflets concept types and XML formatting. Extracted inputs are checked against their source FDA texts: 10% receive human review and the remaining 90% are reviewed by Qwen3-235B.

### B. Base Small Language Models (SLMs) and Baselines

We employ a family of Qwen-2.5-based SLMs as base models for post-training alignment. Specifically, we evaluate three model sizes—0.5B, 3B, and 7B parameters—to examine the impact of model scale on alignment effectiveness. For each size, we consider both base and instruction-tuned variants of the Qwen models to assess the influence of prior instruction tuning on downstream biomedical D2T performance. To investigate whether domain-adapted pretrained models provide a stronger initialization for biomedical tasks, we further compared with Meditron3-Qwen-2.5-7B, a medically fine-tuned variant of Qwen. We compare its performance against both Qwen-2.5-7B and Qwen-2.5-7B-Instruct under identical alignment and evaluation settings. This comparison allows us to assess the relative benefits of medical domain pretraining versus general-purpose and instruction-tuned base models for biomedical narrative generation. We include the original BioLeaflets T5 baseline and standard GPT-5 via Microsoft Azure. GPT-5 uses the same task prompts as SLM training, default reasoning effort, and an output limit of 8,192 tokens; temperature and top-p are not set.

### C. Post-training Alignment Methods

We investigate four representative post-training alignment strategies spanning supervised and preference-based paradigms —SFT, DPO, ORPO, and GRPO—all applied to the same pretrained Qwen-based SLMs for a fair comparison. We assume familiarity with these methods and describe only the settings specific to this study; full experimental hyperparameters are reported in the Experimental Settings section. Several of our runs additionally use prompt loss weighting (PLW) [17], which assigns a fractional loss to prompt tokens rather than masking them entirely; we set the PLW value via the generation-ratio (Rg) heuristic without further tuning, yielding a value of 0.155 for a mean Rg of 0.51 on our training data.

**Supervised fine-tuning (SFT)** serves as our supervised baseline. We apply SFT to Qwen base and instruction-tuned models across all scales and to the medically fine-tuned Meditron3-Qwen-2.5-7B backbone, training each with either the completion-only objective or the PLW variant. In addition, we evaluate a parameter-efficient fine-tuning (PEFT) [21] variant using QLoRA [23] (4-bit quantized LoRA [22]), applied only to SFT with the PLW settings; we do not extend QLoRA to the preference-based methods, focusing there on full-parameter fine-tuning for consistent comparison.

**Direct preference optimization (DPO)** is applied as a two-stage procedure on top of an SFT-initialized model rather than from scratch: we first perform SFT on one half of the training data for two epochs to initialize the model, then apply DPO on the remaining half for three epochs (five epochs total), using the standard sigmoid-based loss.

**Odds ratio preference optimization (ORPO)** unifies the supervised and preference objectives in a single stage. We apply ORPO to Qwen base, instruction-tuned, and biomedical-tuned models under both the completion-only objective and the PLW strategy, using the same preference dataset as DPO and matching the SFT epoch budget for fair comparison.

**Group relative policy optimization (GRPO)** is our reinforcement learning–based method. We adopt the Dr. GRPO objective [19] with token-level rewards, using two task-derived reward functions: (1) METEOR [18] and (2) semantic similarity computed as cosine similarity between paragraph embeddings. For each input we sample 16 candidate responses (rollout size = 16) with sampling temperature 1.0 and top-p 0.95; all remaining GRPO hyperparameters use their default values. To examine the impact of initialization, we consider two strategies: (1) initializing directly from the Qwen-2.5 instruction-tuned models (Qwen-2.5-xB-Inst-GRPO, x = 0.5, 3, or 7), and (2) initializing from Qwen-2.5-xB models that have undergone one epoch of SFT with PLW on the BioLeaflets training set (Qwen-2.5-xB-Inst-SFT-PLW-1epoch-GRPO, x = 0.5, 3, or 7).

### *D. Non-preferred Sample Synthesis*

Preference-based optimization methods such as DPO and ORPO require training data in the form of paired chosen and rejected responses for each input. While gold-standard reference texts naturally serve as the chosen samples, corresponding non-preferred responses are not directly available and must therefore be synthesized. To construct these preference pairs, we adopt a modified rejection sampling strategy. [20] For each structured biomedical input, we prompt the Qwen-2.5-7B-Instruct model to generate five candidate responses. These candidates represent plausible but varying-quality realizations of the same underlying content, generated by a general-purpose language model that is not specialized for the biomedical domain. We then employ the Qwen3-235B-Thinking model as an automated judge to evaluate the generated candidates against the reference. The model selects the most appropriate non-preferred (rejected) response that remains semantically close to the reference while exhibiting deficiencies in clarity, completeness, and/or biomedical domain-specific expression, to provide a preference learning signal.

### *E. Cross-Dataset Generalizability Evaluation*

To assess model generalizability, we conduct a cross-dataset evaluation in which models trained on the BioLeaflets dataset are directly applied to the FDALeaflets (OpenFDA-derived) dataset without any additional fine-tuning or adaptation. Although both datasets address medication leaflet generation from structured drug information, they originate from different regulatory sources—the European Medicines Agency and the U.S. Food and Drug Administration—introducing a realistic distribution shift. We select the best-performing model for each alignment method. For GRPO, Tables IV and V use a checkpoint selected by its best BioLeaflets test-set performance. All selected models are evaluated on FDALeaflets in a zero-shot inference setting. As an external reference, we include GPT-5 as a proprietary large language model baseline on the same structured inputs. Performance is assessed using the same automatic and model-based evaluation metrics as in the in-domain BioLeaflets experiments, enabling direct comparison of in-domain and out-of-domain performance.

### *F. Experimental Settings and Evaluation*

**Experiment Settings** For SFT experiments, all models are trained for five epochs with a learning rate of 1e-5 and a global batch size of 8. For QLoRA-based SFT experiments, we set the LoRA rank and scaling factor (α) to 64 and use a learning rate of 1e-4. In DPO experiments, models are trained for three epochs with a global batch size of 8. Learning rates are set to 1e-5, 7e-6, and 5e-6 for models with 0.5B, 3B, and 7B parameters, respectively. We evaluate β values of 0.1, 0.3, and 0.6, selecting β = 0.3 for DPO based on the best observed BioLeaflets test-set performance. Detailed results for other β values are not reported. For ORPO experiments, we use the same number of training epochs and batch size as in SFT, and the same learning rate settings as in DPO. Like DPO, ORPO uses β = 0.3, selected using test-set performance. In GRPO experiments, models are trained for 10,000 steps (~1.5 epochs), with checkpoints evaluated every 1,000 steps on the BioLeaflets test set to examine performance over training. Figure 1 uses these test-set curves to examine how performance changes with additional RL training. To make the optimizer more responsive to recent gradient updates, we set the Adam optimizer's $\beta^2$ parameter to 0.98. To mitigate training instability, we adopt smaller learning rates of 5e-6, 2e-6, and 1e-6 for models with 0.5B, 3B, and 7B parameters, respectively. Synthetic data generation, SLM training, and Qwen-based evaluation use eight H100 GPUs.

**Evaluation** We employ a combination of automatic and model-based evaluation metrics to assess model performance. For surface-level text overlap, we use n-gram–based metrics including SacreBLEU, ROUGE-L, and METEOR. To capture semantic similarity beyond lexical overlap, we additionally report embedding-based metrics, namely MoverScore and textual semantic similarity (STS). Text embeddings for MoverScore and STS are computed using the ModernBERT (base) model and the Qwen3-Embedding-4B model, respectively. We do not include BERTScore or BLEURT in our evaluation, as both methods are limited to a maximum input length of 512 tokens; in contrast, our evaluation does not apply any truncation, which is necessary to faithfully assess long-form biomedical narratives. For consistency, we re-evaluate results reported in the original BioLeaflets study. Further, we evaluate model outputs using an LLM-as-judge approach (Qwen3-235B-Thinking) across four dimensions: adequacy, hallucination presence, entity inclusion, and fluency. Each dimension is scored on a scale from 1 to 10. Adequacy estimates the overall generation quality by considering fluency, hallucination level, and the completeness of entity inclusion; higher scores indicate better performance for all metrics except hallucination presence, where lower scores indicate fewer hallucinations. These are model-assigned ratings, not measured clinical error rates.

## III. RESULTS

**Table I** summarizes the performance of SFT across model sizes, architectures, and optimization variants. Overall, performance improves monotonically with model scale across all evaluation metrics. Among completion-only training runs, 7B models consistently achieve the strongest results, with Qwen-2.5-7B attaining the highest scores on SacreBLEU, MoverScore, and STS, Qwen-2.5-7B-Instruct the highest ROUGE-L, and the medically fine-tuned Meditron3-Qwen-2.5-7B the highest METEOR. The medically fine-tuned Meditron3-Qwen-2.5-7B model performs comparably to general-purpose Qwen models, particularly on STS, METEOR and ROUGE-L, indicating strong lexical and semantic alignment with reference texts. Applying prompt loss weighting (PLW) yields consistent and substantial gains across all model sizes and architectures. At 7B scale, PLW-enabled models achieve the strongest overall performance, with Qwen-2.5-7B reaching METEOR of 0.6961, ROUGE-L of 0.7103, and STS of 0.9659. Across all configurations, SFT-trained Qwen models generally outperform the baseline T5-BioLeaflet and GPT-5 across most metrics.

The QLoRA results (Qwen-2.5-xB-SFT-PLW-QLoRA) exhibit substantial performance degradation across all metrics, particularly in semantic similarity (STS decreases to 0.90–0.92). These findings suggest that, for medication leaflet generation from structured data, parameter-efficient fine-tuning imposes meaningful limitations compared to full-parameter adaptation due to aggressive parameter constraints.

TABLE I. SUPERVISED FINE-TUNING (SFT) PERFORMANCE ON THE BIOLEAFLETS TEST SET.

| Model Name | METEOR | SacreBleu | Rouge-L | MoverScore | STS |
|---|---|---|---|---|---|
| T5-BioLeaflet | 0.3315 | 50.1962 | 0.4228 | 0.6178 | 0.8975 |
| GPT-5 | 0.4018 | 56.3391 | 0.3799 | 0.6157 | 0.8960 |
| Qwen-2.5-0.5B-SFT | 0.5568 | 47.6332 | 0.5430 | 0.6652 | 0.9424 |
| Qwen-2.5-0.5B-Inst-SFT | 0.5455 | 54.5948 | 0.5348 | 0.6618 | 0.9381 |
| Qwen-2.5-3B-SFT | 0.6280 | 63.4687 | 0.6194 | 0.6952 | 0.9564 |
| Qwen-2.5-3B-Inst-SFT | 0.6219 | 68.7869 | 0.6264 | 0.6928 | 0.9583 |
| Qwen-2.5-7B-SFT | 0.6428 | 69.6013 | 0.6531 | 0.7052 | 0.9605 |
| Qwen-2.5-7B-Inst-SFT | 0.6455 | 66.1887 | 0.6537 | 0.7046 | 0.9585 |
| Meditron3-Qwen-2.5-7B-SFT | 0.6498 | 62.0642 | 0.6498 | 0.7024 | 0.9579 |
| Qwen-2.5-0.5B-SFT-PLW | 0.5914 | 71.4562 | 0.5988 | 0.6791 | 0.9500 |
| Qwen-2.5-0.5B-Inst-SFT-PLW | 0.5948 | 59.2444 | 0.5965 | 0.6761 | 0.9484 |
| Qwen-2.5-3B-SFT-PLW | 0.6597 | 62.4285 | 0.6767 | 0.7079 | 0.9613 |
| Qwen-2.5-3B-Inst-SFT-PLW | 0.6660 | 66.7656 | 0.6816 | 0.7083 | 0.9612 |
| Qwen-2.5-7B-SFT-PLW | **0.6961** | **73.3925** | **0.7103** | **0.7206** | **0.9659** |
| Qwen-2.5-7B-Inst-SFT-PLW | 0.6801 | 71.6954 | 0.6965 | 0.7154 | 0.9625 |
| Meditron3-Qwen-2.5-7B-SFT-PLW | 0.6894 | 67.9088 | 0.7062 | 0.7184 | 0.9646 |
| Qwen-2.5-0.5B-SFT-PLW-QLoRA | 0.3208 | 17.5246 | 0.1952 | 0.5953 | 0.9014 |
| Qwen-2.5-3B-SFT-PLW-QLoRA | 0.3899 | 16.5430 | 0.2555 | 0.6147 | 0.9100 |
| Qwen-2.5-7B-SFT-PLW-QLoRA | 0.4368 | 10.6706 | 0.3679 | 0.6332 | 0.9241 |

**Table II** summarizes performance of models trained with DPO. Compared with SFT (**Table I**), DPO does not yield consistent improvements across metrics or model scales. While some instruction-tuned variants—particularly Qwen-2.5-3B-Instruct and Qwen-2.5-7B-Instruct—achieve competitive or slightly higher scores on selected metrics such as SacreBLEU, ROUGE-L, and STS, these gains are not uniform and are often accompanied by reduced performance in lexical overlap or semantic similarity measures relative to their SFT counterparts. Across model sizes, DPO-trained models generally underperform the best SFT models with prompt loss weighting, especially for larger models where SFT exhibits stronger and more stable gains. The biomedical-tuned Meditron3-Qwen-2.5-7B model also does not demonstrate a clear advantage under DPO compared to SFT. Overall, these results suggest that, in the medication leaflet D2T setting, preference-only optimization via DPO provides limited benefit over supervised fine-tuning and may be less effective at preserving factual and stylistic alignment with reference biomedical narratives.

TABLE II. DIRECT PREFERENCE OPTIMIZATION (DPO) RESULTS ON THE BIOLEAFLETS TEST SET.

| Model Name | METEOR | SacreBleu | Rouge-L | MoverScore | STS |
|---|---|---|---|---|---|
| Qwen-2.5-0.5B-DPO | 0.4543 | 63.2435 | 0.4285 | 0.4879 | 0.9224 |
| Qwen-2.5-0.5B-Inst-DPO | 0.4756 | 62.9512 | 0.4278 | 0.5299 | 0.9307 |
| Qwen-2.5-3B-DPO | 0.4996 | 68.0578 | 0.4700 | 0.6318 | 0.9369 |
| Qwen-2.5-3B-Inst-DPO | 0.5974 | 63.4832 | 0.6264 | 0.6022 | 0.9575 |
| Qwen-2.5-7B-DPO | 0.5963 | 60.6697 | 0.6067 | **0.6689** | 0.9543 |
| Qwen-2.5-7B-Inst-DPO | **0.6226** | **71.5994** | **0.6575** | 0.6371 | **0.9611** |
| Meditron3-Qwen-2.5-7B-DPO | 0.5894 | 68.6461 | 0.5850 | 0.6683 | 0.9526 |

**Table III** summarizes ORPO performance across model scales. ORPO generally improves METEOR, ROUGE-L, MoverScore, and STS over SFT (Table I) and DPO (Table II), although SacreBLEU gains vary by configuration. For example, STS rises from 0.9659 to 0.9740 for Qwen-2.5-7B (+0.0081), and from 0.9646 to 0.9750 for Meditron3-Qwen-2.5-7B (+0.0104). Qwen-2.5-7B-Instruct with ORPO and PLW reaches the highest STS (0.9752), compared with 0.9611 under DPO (+0.0141). The strongest ORPO configurations also achieve METEOR of 0.7676, ROUGE-L of 0.7686, and MoverScore of 0.7633. PLW further improves several metrics, particularly for larger models. These results favor ORPO for reference similarity in this setting.

TABLE III. ORPO Results on the BioLeaflets Test Set

| Model Name | METEOR | SacreBleu | Rouge-L | MoverScore | STS |
|---|---|---|---|---|---|
| Qwen-2.5-0.5B-ORPO | 0.6618 | 58.7732 | 0.6537 | 0.7110 | 0.9579 |
| Qwen-2.5-0.5B-Inst-ORPO | 0.6627 | 62.1891 | 0.6575 | 0.7105 | 0.9568 |
| Qwen-2.5-3B-ORPO | 0.7389 | 62.1689 | 0.7336 | 0.7455 | 0.9700 |
| Qwen-2.5-3B-Inst-ORPO | 0.7347 | 62.2350 | 0.7308 | 0.7454 | 0.9696 |
| Qwen-2.5-7B-ORPO | 0.7510 | 65.2954 | 0.7498 | 0.7555 | 0.9740 |
| Qwen-2.5-7B-Inst-ORPO | 0.7534 | 55.9031 | 0.7507 | 0.7566 | 0.9720 |
| Meditron3-Qwen-2.5-7B-ORPO | 0.7512 | 66.3091 | 0.7488 | 0.7515 | 0.9736 |
| Qwen-2.5-0.5B-ORPO-PLW | 0.6742 | 62.2610 | 0.6633 | 0.7179 | 0.9597 |
| Qwen-2.5-0.5B-Inst-ORPO-PLW | 0.6875 | 64.9656 | 0.6776 | 0.7229 | 0.9617 |
| Qwen-2.5-3B-ORPO-PLW | 0.7513 | 73.1865 | 0.7501 | 0.7544 | 0.9731 |
| Qwen-2.5-3B-Inst-ORPO-PLW | 0.7502 | 62.1489 | 0.7505 | 0.7549 | 0.9702 |
| Qwen-2.5-7B-ORPO-PLW | 0.7631 | 68.0813 | 0.7616 | 0.7618 | 0.9740 |
| Qwen-2.5-7B-Inst-ORPO-PLW | 0.7661 | **73.9364** | 0.7647 | **0.7633** | **0.9752** |
| Meditron3-Qwen-2.5-7B-ORPO-PLW | **0.7676** | 71.0961 | **0.7686** | 0.7631 | 0.9750 |

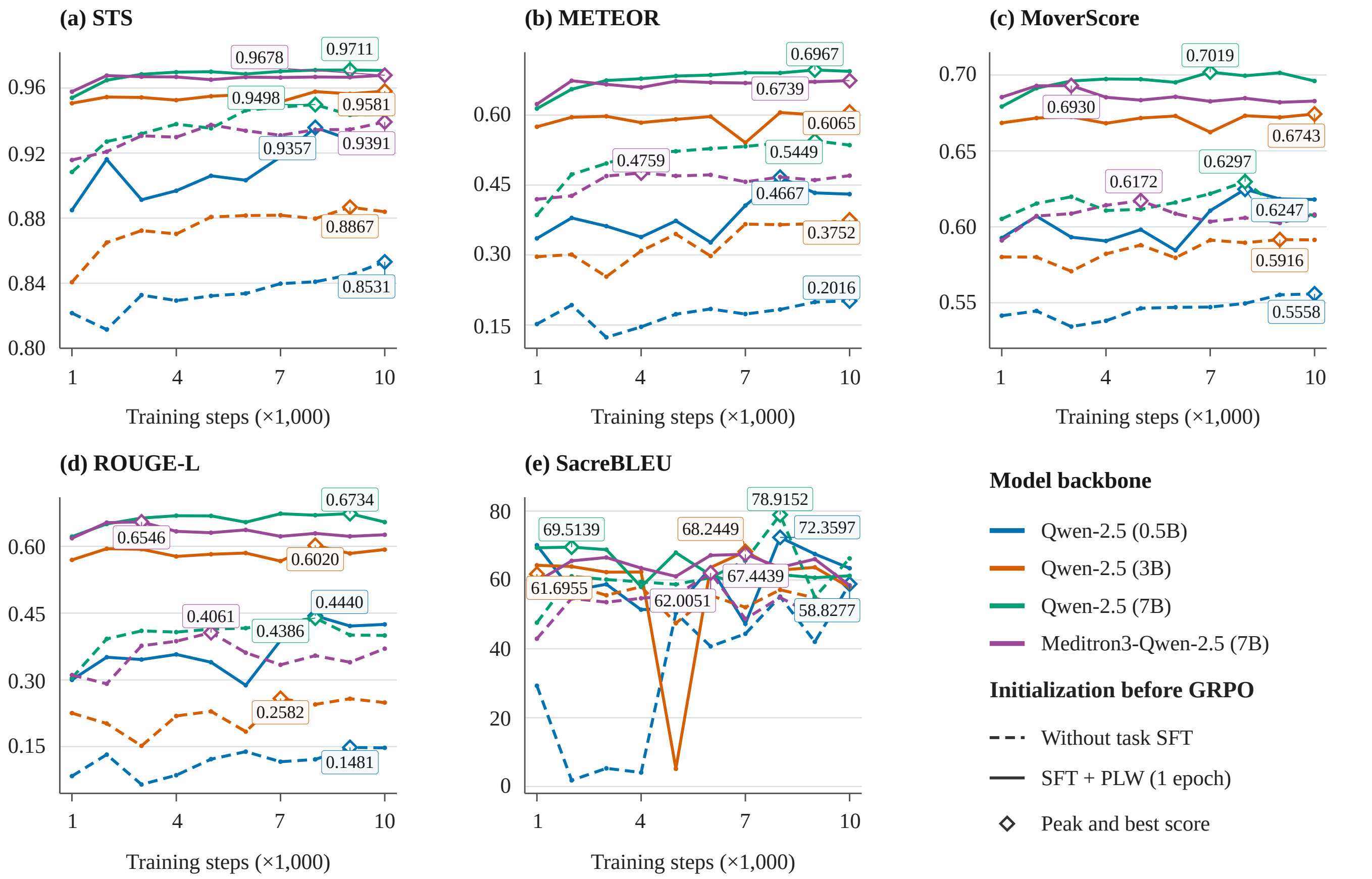

Fig. 1. BioLeaflets test-set performance during GRPO training. Colors denote model backbones; dashed curves start without task SFT, and solid curves start after one epoch of SFT with PLW. Hollow diamonds and numeric labels mark each curve's maximum for that metric; peaks may occur at different checkpoints. Higher scores are better for all metrics.

**Fig. 1** examines test-set performance over RL training; additional steps do not consistently improve every metric. Models initialized from domain-specific SFT checkpoints (Qwen-2.5-xB-Inst-SFT-PLW-1epoch-GRPO) generally achieve higher lexical and semantic scores than those initialized directly from instruction-tuned models (Qwen-2.5-xB-Inst-GRPO), while SacreBLEU fluctuates substantially. Among all GRPO models, the Qwen-2.5-7B-Inst-SFT-PLW-1epoch-GRPO configuration emerged as the most robust, achieving a peak semantic similarity (STS) score of 0.9711 and a METEOR score of 0.6967 at 9,000 steps. Although the instruction-initialized Qwen-2.5-7B-Inst-GRPO model exhibits steady improvements during training, it does not close the performance gap relative to SFT-initialized counterparts. In addition, substituting the base Qwen-2.5-7B backbone with a medical pretrained Meditron3-Qwen-2.5-7B does not yield a measurable advantage, with both backbones achieving comparable evaluation scores across METEOR and STS. We also observe that while smaller models (0.5B and 3B) exhibited measurable gains during training, their absolute performance remained substantially lower than 7B counterparts, particularly on semantic metrics. For example, the best-performing Qwen-2.5-0.5B-Inst-SFT-PLW-1epoch-GRPO model only reaches STS of 0.9357, compared to STS > 0.97 for the 7B variant. This gap suggests that, even with advanced alignment strategies, sufficient model capacity remains critical for capturing nuanced biomedical semantics in D2T generation.

**Table IV** reports the cross-dataset performance on the FDALeaflets test set, where models trained exclusively on the BioLeaflets (EMA-derived) dataset are evaluated zero-shot on the FDA-sourced medication leaflets. The selected aligned SLMs exhibit lower semantic similarity than in-domain evaluation, reflecting the distribution shift induced by differences in regulatory structure, section taxonomy, and linguistic conventions. Among alignment strategies, GRPO demonstrates the strongest generalization behavior. The Qwen-2.5-7B-Inst-GRPO model achieves the highest semantic similarity (0.9407) and METEOR (0.5400), exceeding GPT-5 (STS of 0.9295; METEOR of 0.5328) and models trained with SFT, DPO, or ORPO. Compared with SFT, GRPO improves STS by +0.055 absolute (0.9407 vs. 0.8858), indicating improved semantic robustness under regulatory domain shift. ORPO also generalizes better than SFT and DPO, achieving STS of 0.9179, but remains notably below GRPO. Interestingly, while the SFT-initialized GRPO model performs best in-domain, its zero-shot FDA performance (STS of 0.9023) trails that of instruction-initialized GRPO, suggesting a trade-off between domain specialization and cross-regulatory adaptability. Taken together, these results indicate that reinforcement learning-based alignment, particularly GRPO, yields more robust semantic generalization than supervised or pairwise preference-based methods when transferring biomedical D2T models across regulatory sources.

TABLE IV. CROSS-DATASET EVALUATION ON THE FDALEAFLETS TEST SET.

| Model Name | METEOR | SacreBleu | Rouge-L | MoverScore | STS |
|---|---|---|---|---|---|
| GPT-5 | 0.5328 | 56.8498 | 0.5206 | 0.6412 | 0.9295 |
| Qwen-2.5-7B-SFT-PLW | 0.4416 | 50.3102 | 0.4391 | 0.6141 | 0.8858 |
| Qwen-2.5-7B-Inst-DPO | 0.4359 | 50.8516 | 0.4558 | 0.6300 | 0.8910 |
| Qwen-2.5-7B-Inst-ORPO-PLW | 0.5147 | 55.2021 | 0.5281 | 0.6405 | 0.9179 |
| Qwen-2.5-7B-Inst-GRPO | **0.5400** | **70.3194** | **0.5314** | **0.6436** | **0.9407** |
| Qwen-2.5-7B-Inst-SFT-PLW-1epoch-GRPO | 0.5083 | 50.6604 | 0.4997 | 0.6268 | 0.9023 |

TABLE V. LLM-AS-JUDGE EVALUATION RESULTS ON THE BIOLEAFLETS AND OPENFDA TEST SETS.

| Model Name | Dataset | Hallucination * | Entity Inclusion* | Fluency* | Adequacy* |
|---|---|---|---|---|---|
| Qwen-2.5-7B-SFT-PLW | BioLeaflets | 5.6(2.7) | 6.7(2.3) | 7.5(1.5) | 5.7(1.9) |
| Qwen-2.5-7B-Inst-DPO | | 5.6(2.0) | 6.1(2.3) | 6.5(2.0) | 4.8(1.6) |
| Qwen-2.5-7B-Inst-ORPO-PLW | | 4.9(3.1) | 7.5(2.3) | 8.1(1.4) | **6.3(2.1)** |
| Qwen-2.5-7B-Inst-GRPO | | 5.3(1.2) | 6.7(1.7) | 8.5(1.2) | 5.6(1.2) |
| Qwen-2.5-7B-Inst-SFT-PLW-1epoch-GRPO | | 5.1(2.2) | **7.8(2.1)** | 8.5(1.4) | **6.3(1.5)** |
| GPT-5 | | **4.7(2.4)** | 6.5(1.8) | **8.7(1.2)** | 5.9(1.2) |
| Qwen-2.5-7B-SFT-PLW | OpenFDA | 8.6(1.4) | 4.6(1.8) | 5.0(1.7) | 3.6(1.3) |
| Qwen-2.5-7B-Inst-DPO | | 8.3(2.0) | 4.8(2.1) | 4.0(2.0) | 3.6(1.6) |
| Qwen-2.5-7B-Inst-ORPO-PLW | | 5.9(2.0) | 6.1(1.8) | 7.1(1.7) | 5.9(1.3) |
| Qwen-2.5-7B-Inst-GRPO | | 5.8(1.9) | 6.7(1.9) | **7.4(1.6)** | **6.0(1.2)** |
| Qwen-2.5-7B-Inst-SFT-PLW-1epoch-GRPO | | 5.4(1.6) | 6.4(1.8) | 7.4(1.6) | 6.0(1.3) |
| GPT-5 | | **4.6(2.7)** | **6.8(1.6)** | **7.7(1.3)** | **6.2(1.2)** |

a. *Values on a scale from one to ten; average and standard deviation are reported.

**Table V** reports LLM-as-judge ratings on BioLeaflets and FDALeaflets across adequacy, hallucination presence, entity inclusion, and fluency. On BioLeaflets, SFT-initialized GRPO achieves the highest entity inclusion (7.8) and ties for the highest adequacy (6.3), while GPT-5 has the highest fluency (8.7) and lowest hallucination score (4.7). On FDALeaflets, SFT and DPO receive high hallucination scores (8.6 and 8.3) and low adequacy (3.6). Both GRPO variants achieve adequacy of 6.0 and fluency of 7.4, improving on these SLM baselines. However, GPT-5 achieves lower hallucination (4.6) and higher adequacy (6.2) than either GRPO variant. Thus, GRPO's stronger reference similarity in Table IV does not establish superior factual accuracy or clinical reliability. The metrics capture different aspects of generation, and their rankings should be interpreted together. These descriptive differences do not establish statistical significance.

## IV. DISCUSSION

This study systematically evaluates post-training alignment strategies for SLMs in medication leaflet generation from structured information. Our findings demonstrate that advanced alignment strategies beyond standard SFT emerge as dominant factors influencing semantic fidelity, robustness, and generalizability, often exceeding the impact of model size or pretraining domain.

Among reinforcement learning-free methods, ORPO consistently achieves the best performance on both semantic and lexical metrics across all model sizes. By jointly optimizing likelihood and preference signals, ORPO preserves strong supervision while enabling preference-based refinement, which appears particularly well suited to highly constrained biomedical generation tasks requiring close adherence to source content while improving stylistic and semantic alignment. In contrast, DPO underperforms both ORPO and the SFT baseline. Recent work has identified a phenomenon known as DPO *likelihood* *displacement*, [24] where the probabilities of both preferred and rejected outputs can decrease during training, leading to erosion of supervised signals and degradation of semantic fidelity when preference pairs do not exhibit sufficiently distinct differences. This sensitivity to the quality and contrastiveness of preference data implies that collecting or synthesizing high contrast rejected samples that are both meaningfully inferior to preferred outputs yet not trivially incorrect poses a significant challenge for DPO. Our results further indicate that although ORPO can effectively leverage rejected samples, DPO struggles to learn a stable and reliable preference margin without destabilizing the underlying supervised signal, despite both methods being trained on the same preference dataset. These findings highlight that constructing high-quality preference data for DPO remains an open and nontrivial challenge, particularly for biomedical text generation tasks such as medication leaflet generation.

Within SFT, parameter-efficient fine-tuning via QLoRA yields substantially worse performance than full-parameter adaptation. This degradation suggests that the reduced number of trainable parameters limits the model's capacity to learn the complex mappings between structured inputs and coherent narratives. For medication leaflet generation, full-parameter adaptation remains critical to achieving high semantic fidelity. We also observe that Prompt Loss Weighting (PLW) consistently improves performance in both SFT and ORPO settings. By assigning a fractional loss to prompt tokens rather than masking them entirely as standard instruction fine-tuning, PLW provides additional gradient signal for tasks with long, information-dense instructions, which is a common characteristic of biomedical D2T generation. These findings support PLW as a simple yet effective training strategy for structured biomedical generation tasks combined with SFT or ORPO.

We next examine GRPO, a reinforcement learning-based alignment method that optimizes relative preferences among groups of generated outputs. GRPO demonstrates competitive performance but is highly sensitive to initialization. Models initialized from domain-specific SFT checkpoints achieve stronger in-domain results, indicating that reinforcement learning benefits from a warm start within the target data manifold. Reward design also plays a critical role. We evaluate multiple reward formulations based on lexical and semantic metrics and find that combining METEOR and semantic textual similarity (STS) yields the most stable improvements. However, GRPO remains challenging for generation tasks, as reward signals derived from automatic metrics are inherently indirect and noisier than the explicit correctness signals available in domains such as mathematics or code generation. Because METEOR and semantic similarity also contribute to evaluation, gains on these metrics may partly reflect optimization of the evaluation criteria and do not independently establish factual accuracy. HERO similarly investigates reward granularity and reasoning–diagnosis consistency for reducing hallucinations in radiology report generation. [25]

Across all alignment strategies, the medical pretrained Qwen-2.5-7B model, Meditron3-Qwen-2.5-7B, does not outperform its non-medical Qwen-2.5-7B counterpart. This suggests that medication leaflet generation is a highly specialized task, and that broad medical-domain pretraining alone may be insufficient. Future work may require pretraining on medication- or regulatory-specific corpora to realize gains from domain-adaptive pretraining.

Full-parameter Qwen models generally outperform the re-evaluated T5 baseline on semantic similarity. Differences in architecture, scale, pretraining, and fine-tuning may contribute; this study does not isolate their individual effects.

Selected fine-tuned SLMs exceed GPT-5 on several lexical and semantic similarity metrics. However, GPT-5 has better FDALeaflets hallucination and adequacy ratings, precluding a general claim of SLM superiority. Training SLMs enables direct task-specific optimization, whereas our GPT-5 comparison uses prompting. Notably, although ORPO and SFT-initialized GRPO achieve better in-domain performance, the Qwen-2.5-7B-Inst-GRPO, initialized without SFT, exhibits the strongest cross-dataset generalization. This suggests that while supervised training/initialization promotes specialization to a specific regulatory framework, it may cause overfitting to dataset-specific conventions. In contrast, the GRPO model trained without intermediate SFT appears to learn more flexible generation strategies guided by semantic reward signals, resulting in superior robustness under regulatory distribution shift.

This study has limitations. DPO/ORPO $\beta$ tuning and GRPO checkpoint selection use BioLeaflets test scores, so in-domain estimates may be optimistic; Figure 1's curves should be interpreted descriptively. Qwen3-235B-Thinking constructs FDALeaflets inputs, selects rejected preference responses, and judges outputs. Reusing one model across these roles may introduce correlated biases. FDALeaflets contains errors in the extracted entities, which may affect cross-dataset findings. Human review covers 10% of inputs; Qwen review of the remainder is not independent validation. Future work should use a more rigorous LLM review process or human annotation of the entire dataset. LLM ratings complement reference-based metrics but are not independent clinical validation. We do not conduct human evaluation of generated leaflets for factuality, safety, or patient comprehension. Before patient-facing use, qualified clinicians should verify generated content against source labels, including dosing, contraindications, and adverse events; intended users should assess readability and comprehension. Our findings are limited to medication leaflets and require independent evaluation across other biomedical D2T tasks.

## V. Conclusions

This study compares post-training alignment strategies for small language models in medication leaflet generation. ORPO achieves the strongest in-domain semantic performance among RL-free methods, while GRPO achieves the highest FDALeaflets reference-similarity scores among evaluated models. Full-parameter fine-tuning and prompt loss weighting improve performance in the tested settings. Selected SLMs exceed GPT-5 on reference-based metrics, but GPT-5 obtains better FDALeaflets hallucination and adequacy ratings. These findings support further study of task-specific alignment; they do not establish clinical readiness or factual superiority.

Independent evaluation and human verification remain necessary before patient-facing use.

## Data and Source Code Availability

Code and data are available in the GitHub repository.